\documentclass[11pt,a4paper]{article}
\usepackage[hyperref]{acl2019}
\usepackage{times}
\usepackage{latexsym}
\usepackage{graphicx}
\usepackage{booktabs}

\usepackage{url}

\aclfinalcopy 
\def\aclpaperid{1761} 

\newcommand{\Sref}[1]{\S\ref{#1}}

\newcommand{\Fref}[1]{Figure~\ref{#1}}

\newcommand{\Tref}[1]{Table~\ref{#1}}

\newcommand{\ignore}[1]{}

\title{Entity-Centric Contextual Affective Analysis}

\author{Anjalie Field \\
  Carnegie Mellon University \\
  \texttt{anjalief@cs.cmu.edu} \\\And
  Yulia Tsvetkov \\
  Carnegie Mellon University \\
  \texttt{ytsvetko@cs.cmu.edu} \\}

\date{}

\begin{document}
\maketitle
\begin{abstract}
While contextualized word representations have improved state-of-the-art benchmarks in many NLP tasks, their potential usefulness for social-oriented tasks remains largely unexplored. We show how contextualized word embeddings can be used to capture affect dimensions in portrayals of people. We evaluate our methodology quantitatively, on held-out affect lexicons, and qualitatively, through case examples. We find that contextualized word representations do encode meaningful affect information, but they are heavily biased towards their training data, which limits their usefulness to in-domain analyses. We ultimately use our method to examine differences in portrayals of men and women.
\end{abstract}

\section{Introduction}

Pre-trained contextualized word embeddings \cite{peters2018deep, devlin2018bert, Radford18} have become increasingly common in natural language processing (NLP), improving state-of-the-art results in many standard NLP tasks. However, beyond standard tasks, NLP tools are also vital to more open-ended exploratory tasks, particularly in social science.
How these types of tasks can benefit from pre-trained contextualized embeddings has not yet been explored.

In this work, we show how to leverage these embeddings to conduct \emph{entity-centric} analyses, which broadly seek to address how entities are portrayed in narrative text \cite{bamman2013learning, card2016analyzing}. For instance, in the sentence ``Batman apprehends the Joker'', a reader might infer that Batman is good, the Joker is evil, and Batman is more powerful than the Joker. Analyzing how people are portrayed in narratives is a key starting point to identifying stereotypes and bias \cite{joseph2017girls, fast2016shirtless, field2019}. 

Existing methods for analyzing people portrayals take either an unsupervised approach \cite{bamman2013learning}, which requires large amounts of data and can be difficult to interpret, or rely on domain-specific knowledge \cite{fast2016shirtless, wagner2015s}, which does not generalize well to other hypotheses and data domains. Furthermore, most models are limited to discrete word-level features, whereas continuous-valued embeddings are typically more expressive. We introduce a novel approach to analyzing entities that maps contextualized embeddings to interpretable dimensions.

Specifically, we propose using pre-trained embeddings to extract \emph{affect} information about target entities. Social psychology research has identified 3 primary affect dimensions:  \emph{Potency} (strength/weakness of an identity), \emph{Valence} (goodness/badness of an identity), and \emph{Activity} (activeness/passiveness of an identity) \cite{osgood1957measurement, russell1980circumplex, russell2003core}. We refer to these dimensions as \textbf{power}, \textbf{sentiment}, and \textbf{agency} for consistency with prior work in NLP \cite{SapConnotationFilms, RashkinConnotationInvestigation, field2019}. Thus, in the previous example, ``Batman apprehends the Joker'', we might associate Batman with high power, high sentiment, and high agency. 

While much literature in NLP has examined sentiment, analyses of power have largely been limited to a dialog setting \cite{prabhakaran2015social}, and almost no work has examined agency. We propose that mapping entities into these 3 dimensions provides a framework for examining narratives that is more holistic than sentiment analyses and more generalizable than task-specific frameworks. The idea that these 3 dimensions are sufficient for capturing affect has also formed the basis of social psychological models \cite{heise2007expressive, alhothali2015good}. 

Drawing from this theory, we combine contextualized word embeddings with affect lexicons \cite{vad-acl2018} to obtain power, sentiment, and agency scores for entities in narrative text. After describing our methodology (\Sref{sec:methods}), we evaluate how well these contextualized embeddings capture affect information on held-out lexicons (\Sref{sec:lex_corr}). We then evaluate how well our method scores entities on manually curated benchmarks (\Sref{sec:quant_entity}) and through qualitative examples (\Sref{sec:qual_entity}). Finally, we use our method to examine different portrayals of men and women (\Sref{sec:usage}), focusing on the same domains as prior work \cite{wagner2015s, fu2016tie}. Ultimately, our work suggests that contexualized embeddings have the potential to improve analyses of entity portrayals. However, we find that these representations are biased towards portrayals in the training data, which limits their usefulness to analyzing in-domain data.

Our contributions in this work include: (1) a novel method for analyzing entities in a narrative that is both interpretable and generalizable, (2) an assessment of how well contextualized word embeddings capture affect information, and (3) an analysis of entity portrayals in various domains.

\section{Methodology}
\label{sec:methods}

Given an entity, such as ``Batman'', mentioned in a narrative, our goal is to obtain power, sentiment, and agency scores for the entity. We take two approaches: supervised regression and semi-supervised embedding projection. For both approaches, we use pre-trained contextualized embeddings as features and we use the NRC Valence, Arousal, and Dominance (VAD) Lexicon as training and test data \cite{vad-acl2018}. While we use this lexicon because its annotations contain our target dimensions of power, sentiment, and agency, our methodology readily generalizes to other lexicons.

\subsection{Regression Model}

In the regression model, we take a supervised approach, using annotations from the NRC VAD Lexicon as training data. 

Given a training word $w$ and a large training corpus, we extract a contextual embedding $\mathbf{e}$ for every instance of $w$ in the corpus. We use off-the-shelf pre-trained language models to extract sentence-level embeddings with no additional fine-tuning. Then, we average over all $\mathbf{e}$ embeddings for each instance $w$ to obtain a single feature vector for each training point. We then train a Kernel Ridge Regression model using these embeddings as features.\footnote{We also experimented with Linear Regression and Ridge Regression, but found that Kernel Ridge Regression performed the best.}

To extract affect scores for an entity in a narrative, we use the same pre-trained language model to extract a contextual embedding for the entity.  Then, we feed this embedding through the regression model to obtain power, sentiment, and agency scores. When an entity occurs multiple times in the narrative, we average over the contextual embeddings for each occurrence of the entity and score the averaged embedding.

\begin{table}
\begin{tabular}{lll}
& \textbf{Low} & \textbf{High} \\
& timid              & resourceful         \\
& weakly             & powerfully          \\
 Power & cowardly           & courageous          \\
& inferior           & superior            \\
& clumsy             & skillful            \\
\hline
& negative              & positive         \\
& pessimistic             & optimistic          \\
Sentiment & annoyed           & amused          \\
& pessimism           & optimism            \\
& disappointed             & pleased      \\
\hline
& silently              & furiously         \\
& meek             & lusty          \\
Agency & homely           & sexy          \\
& bored           & flustered            \\
& quietly             & frantically
\end{tabular}
\caption{Polar-opposite word pairs identified by ASP}
\label{tab:power_pairs}
\end{table}

\subsection{Affect Subspace Projection (ASP)}

The main disadvantage of the regression approach is that we are unable to control for confounds and prevent overfitting to the training data. For example, many low-agency nouns tend to be inanimate objects (i.e. \textit{table}), while high-agency nouns are people-oriented words (i.e. \textit{dictator}). Thus, we can expect that the model learns to predict the difference between  classes of nouns, rather than solely learning the affect dimension of interest. While other variations of regression allow for the inclusion of covariates and confounds, we have no systematic way to quantify or even identify these confounds. Instead, we devise a method to isolate dimensions of power, agency, and sentiment by first identifying corresponding subspaces in the embedding space and then projecting entities onto these dimensions. We refer to this method as \textit{affect subspace projection} (ASP).

We describe this process for obtaining power scores; the agency and sentiment dimensions are analogous. In order to isolate the power subspace, we draw inspiration from \cite{bolukbasi2016man}. First, we need to identify pairs of words whose meanings differ only in that one word connotes high power and the second word connotes low power. We define a set $\mathcal{H}$, which consists of the $|\mathcal{H}|$ highest-powered words from the VAD lexicon and a set $\mathcal{L}$, which consists of the $|\mathcal{L}|$ lowest powered words from the VAD Lexicon. For every word $w_h \in \mathcal{H}$, we use cosine similarity over contextual embedding representations to identify $w_l \in \mathcal{L}$, the low-powered word that is most similar to $w_h$. We allow each $w_l$ to match to at most one $w_h$. Thus, we identify pairs of words ($w_h$, $w_l$), where $w_h$ and $w_l$ are very similar words but with polar opposite power scores. Finally, we keep only the $N$ pairs with the greatest cosine similarity. We tune hyperparameters $|\mathcal{H}|$, $|\mathcal{L}|$, and $N$ over a validation set. We show examples of extracted pairs for each dimension in \Tref{tab:power_pairs}.

Next, we use these paired words to construct a set of vectors whose direction of greatest variance is along the power subspace.
For each pair of high and low power words ($w_h$, $w_l$), we take their embedding representations $\mathbf{e_h}$ and $\mathbf{e_l}$ in the same way as in the regression model.
We then define $\mathbf{\mu} = (\mathbf{e_h} + \mathbf{e_l}) / 2$, and construct a matrix $\mathbf{M}$, where each row is $\mathbf{e_l} - \mathbf{\mu}$ or $\mathbf{e_h} - \mathbf{\mu}$. 
Thus, $\mathbf{M}$ is a $d \times 2N$ dimensional matrix, where $d$ is the dimension of the embeddings.
We then run PCA over $\mathbf{M}$ to extract its principle components.
For all 3 affect dimensions, the first principle component captures the highest percentage of variance (Appendix \ref{sec:appendixA}), followed by a sharp drop off.
Thus, we keep the first principle component as the target subspace.

Finally, to score an entity in a narrative, we take the entity's contextual embedding representation and project it onto the identified subspace.
Because we keep only the first principle component as the target subspace, the projection results in a single-dimensional vector, i.e., a power score. 
We repeat the process for agency and sentiment, constructing 3 separate $\mathbf{M}$ matrices in order to obtain power, sentiment, and agency scores.

\section{Experimental Setup}

The NRC VAD Lexicon contains valence (sentiment), arousal (agency), and dominance (power) annotations for more than 20,000 English words. It was created through manual annotations using Best--Worst scaling. The final annotations are on a scale from 0 (i.e. lower power) to 1 (i.e. high power) \cite{vad-acl2018}. We randomly divide the lexicon into training (16,007), dev (2,000), and test (2,000) sets. 

We extract embeddings to train our models from a corpus of 42,306 Wikipedia movie plot summaries \cite{bamman2013learning}.\footnote{When experimenting with other training corpora, such as newspaper articles, we found the choice of training corpus had little impact on results.} We use two pretrained language models to extract embeddings: ELMo \cite{peters2018deep} and BERT \cite{devlin2018bert}. It is important to note that the movie plots corpus we used for extraction is distinct from the corpora used to train ELMo (5.5B tokens from Wikipedia and WMT news crawl) and BERT (800M-word BooksCorpus and 2,500M-word Wikipedia).

We use two variants of BERT to extract embeddings. In the first, referred to as ``BERT-masked'', we mask out the target word before extracting embeddings from an input sentence. Masking out target words is a part of the BERT training objective \cite{devlin2018bert}. By using masks in our embedding extractions, we force the model to produce an embedding solely from the context surrounding the word, rather than relying on information from the word itself. In the second variant, referred to as ``BERT'',  we extract embeddings over each sentence containing a target without modification. We report further details including hyperparamter settings in Appendix \ref{sec:appendix}.

\begin{table}
\begin{tabular}{lccc}
\toprule
\textbf{Regression}\\
\midrule
& Power & Sentiment & Agency \\
ELMo & 0.78 & \textbf{0.84} & 0.76  \\
BERT & \textbf{0.79 } & 0.83 & \textbf{0.78}\\
BERT-masked & 0.64 & 0.70  &  0.62\\
\\
\toprule
\textbf{ASP} \\
\midrule
& Power & Sentiment & Agency \\
ELMo & 0.65 & 0.76 & 0.63 \\
BERT & 0.65 & 0.71  & 0.66 \\
BERT-masked & 0.41 & 0.47 & 0.41 \\

\end{tabular}
\caption{Pearson correlations between gold NRC VAD labels and scores predicted by our models. Correlations are generally high, with the regression method outperforming ASP. All correlations are statistically significant ($p < 1e-75$).}
\label{tab:pearson_corrs}
\end{table}

\section{Results and Analysis}

\subsection{Lexicon Correlations}
\label{sec:lex_corr}

\Tref{tab:pearson_corrs} shows the Pearson correlations between gold annotations and the scores predicted by our models over the held-out VAD test set.
The high correlations demonstrate that both the regression and ASP models successfully capture information about power, sentiment, and agency from contextualized embeddings.
The ELMo embeddings and unmasked BERT embeddings perform approximately the same.
However, the masked BERT embeddings perform markedly worse than the unmasked embeddings.\footnote{One of the drawbacks of context-based word embeddings is that antonyms like ``positive'' and ``negative'' tend to have similar embeddings, because they tend to be used in similar contexts. However, given the breadth of words in the VAD lexicon, we do expect context to differ for oppositely scored words. For instance we would expect ``pauper'' and ``king'' to be used in different contexts, as well as ``pauper'' and ``powerful''.}
The poorer performance of the masked embeddings demonstrates the extent to which the BERT model biases representations towards the actual observed word, which is explicitly one of the motivations of the BERT training objective \cite{devlin2018bert}.
More specifically, when we mask out the target before extracting embeddings, we force the extracted embedding to only encode information from the surrounding context.
Then any improvements in performance when we do not mask out the target are presumably obtained from the word-form for the target itself.
For example, we may score ``king'' as high-powered because ``king'' often occurred as a high-powered entity in the data used to train the BERT model, regardless of whether or not it appeared to be high-powered in the corpus we ultimately extract embeddings from.
Nevertheless, training with BERT-masked embeddings still results in statistically significant correlations, which suggests that some affect information is derived from surrounding context.

The regression model generally outperforms ASP on this task. The regression model has an advantage over ASP in that it is directly trained over the full lexicon, whereas ASP chooses a subset of extreme words to guide the model. However, as discussed in \Sref{sec:methods}, it is difficult to determine what effect other confounds have on the regression model, while the ASP approach provides more concrete evidence that these contextualized word embeddings encode affect information.

\begin{table}
\begin{tabular}{lcc}

& Regression & ASP \\
ELMo & 0.51 & 0.21 \\
BERT & 0.38 & 0.38 \\
BERT-masked & 0.17 & -0.085 \\
ELMo + Freq & \textbf{0.65} & 0.48 \\
\midrule
Frequency Baseline & \multicolumn{2}{c}{0.61} \\
\citet{field2019} & \multicolumn{2}{c}{-0.12}  \\

\end{tabular}
\caption{Spearman correlations between automatically induced power scores and Forbes power ranking. Correlations for ELMo regression  ($p = 0.029$),  ELMo regression + Freq  ($p = 0.003$), and the frequency baseline ($p = 0.007$) are statistically significant. The ELMo regression + Freq model performs the best.}
\label{tab:forbes_power}
\end{table}

\subsection{Quantitative Analysis of Entity Scores}
\label{sec:quant_entity}

Next, we evaluate how well our models capture affect information in entities, rather than words, by assessing power scores through two metrics. We compare our models against the entity-scoring metric proposed by \citet{field2019} and against a frequency baseline, where we consider an entity's power score to be the number of times the entity is mentioned in the text.

First, we consider an in-domain task, where we compare our metrics for scoring power with a standard benchmark that we expect to be reflected in both the data we use to extract embeddings and the data used to train ELMo and BERT.
More specifically, we use the power scores obtained from our model to rank the 20 most powerful people in 2016 according to Forbes Magazine.\footnote{\url{http://bit.ly/2W5Jvnf}} 

This is a particularly difficult task:  unlike prior work, which seeks to identify the most powerful people in a corpus \cite{field2019}, we seek to rank these people according to their power, which requires more precise scores. 
Furthermore, the frequency metric supplies a particularly strong baseline.
The metrics that Forbes Magazine uses to compose the list of powerful people include a person's influence as well as how actively they use their power.\footnote{\url{http://bit.ly/2Mp2R70}}
Under these conditions, Forbes Magazine may consider a person to be powerful simply because they are mentioned frequently in the media.
Additionally, we can surmise that people who actively use their power are mentioned frequently in the media.

\Tref{tab:forbes_power} presents Spearman correlations between our scores and rank on the Forbes list for each model. For all metrics, we construct embeddings from every instance of each person's full name in U.S.~articles from 2016 in the NOW news corpus.\footnote{\url{https://corpus.byu.edu/now/}} 

In addition to the proposed methods, we used our best performing model (regression with ELMo) to augment the frequency baseline, by normalizing and summing the frequency scores with the scores from this model. This combined model achieves the strongest correlation (raw scores from this model are shown in \Fref{fig:forbes_power}).
Furthermore, the regression with ELMo model alone achieves a statistically significant correlation even without the incorporation of frequency scores. The unmasked BERT embeddings also achieve positively correlated scores, though these correlations are not statistically significant. The BERT-masked embeddings perform particularly poorly, as does the method for scoring power proposed in \citet{field2019}. While \citet{field2019} may be capable of identifying powerful entities, we suspect it is not fine-grained enough to rank them. 

While frequency serves as a strong baseline for power, we would not expect frequency to be a good measure of sentiment or agency. None of our metrics for these traits are significantly correlated with the Forbes' ranking. Also, we would not expect frequency to be a good measure in other contexts, such as how powerfully an entity is portrayed in a single document rather than across a large media corpus.

\begin{table}
\begin{tabular}{lcc}

\toprule
\multicolumn{3}{l}{\textbf{Full annotation set (383 pairs)}} \\
\midrule

& Regression & ASP \\
ELMo &  44.9 & 43.6 \\
BERT & 41.8 & 49.3 \\
BERT-masked & 49.6 & \textbf{59.0} \\
\midrule
Frequency Baseline & \multicolumn{2}{c}{58.0} \\

\toprule
\multicolumn{3}{l}{\textbf{Reduced annotation set (49 pairs)}} \\
\midrule

& Regression & ASP \\
ELMo & 36.7  & 42.8 \\
BERT & 42.9 & 49.0 \\
BERT-masked & 53.1 & 55.1 \\
\midrule
Frequency Baseline & \multicolumn{2}{c}{57.1} \\
\citet{field2019} & \multicolumn{2}{c}{\textbf{71.4}}  \\

\end{tabular}
\caption{Accuracy for scoring how powerful entities are as compared with annotations over articles related to the \#MeToo movement. Our metrics do not consistently outperform the baselines, suggesting ELMo and BERT embeddings fail to transfer across domains.}
\label{tab:metoo_power}
\end{table}

Next, we further explore performance on an out-of-domain task: specifically how powerfully entities are portrayed in a specific set of articles, which we do not expect to align with portrayals in the data used to train ELMo and BERT.

For this task, we use the same evaluation metrics as \citet{field2019}; we compare our scores with hand-annotated power rankings over a set of newspaper articles related to a specific event in the \#MeToo movement, namely allegations of sexual harassment against the comedian Aziz Ansari. Following \citet{field2019}, we interpret the hand-annotations, in which human annotators rank entities according to how powerful they seem, as a pairwise task (is entity A more powerful than entity B?) and compute accuracy over pairs of entities. We discard annotations where annotators strongly disagreed about the power of the entity (i.e. annotations differ by more than 2 ranks).

\citet{field2019} compare results with off-the-shelf connotation frame lexicons, which restricts analysis to a limited set of pairs, since only entities used with verbs from the lexicon are included. In contrast, we simply use string matching to identify entities in the text, without requiring that the entities be linked to specific verbs, allowing for the identification of more entities.

\Tref{tab:metoo_power} shows results over the same set of pairs used for evaluation in \citet{field2019} as well as an expanded set, when we do not restrict to entities used with lexicon verbs.
Our metrics fail to consistently outperform even the frequency baseline for this task, likely because the ELMo and BERT embeddings are biased towards their training data. 

The \#MeToo movement is widely known for subverting traditional power roles: allegations made by traditionally unpowerful women have brought down traditionally powerful men.
For example, Harvey Weinstein, an influential film producer, has traditionally been a powerful figure in society, but numerous allegations of sexual harassment have resulted in his effective removal from the industry. While articles about the \#MeToo movement portray men like Weinstein as unpowerful, we can speculate that the corpora used to train ELMo and BERT portray them as powerful.

Thus, in a corpus where traditional power roles have been inverted, the embeddings extracted from ELMo and BERT perform worse than random, as they are biased towards the power structures in the data they are trained on. Further evidence of this exists in the performance of the BERT-masked embeddings - whereas these embeddings generally capture power poorly as compared to the unmasked embeddings (\Tref{tab:pearson_corrs}), they outperform the unmasked embeddings on this task, and even outperform the frequency baseline in one setting. Nevertheless, they do not outperform \citet{field2019}, likely because they do not capture affect information as well as the unmasked embeddings (\Tref{tab:pearson_corrs}).

\begin{figure}
    \centering
    \includegraphics[width=\linewidth]{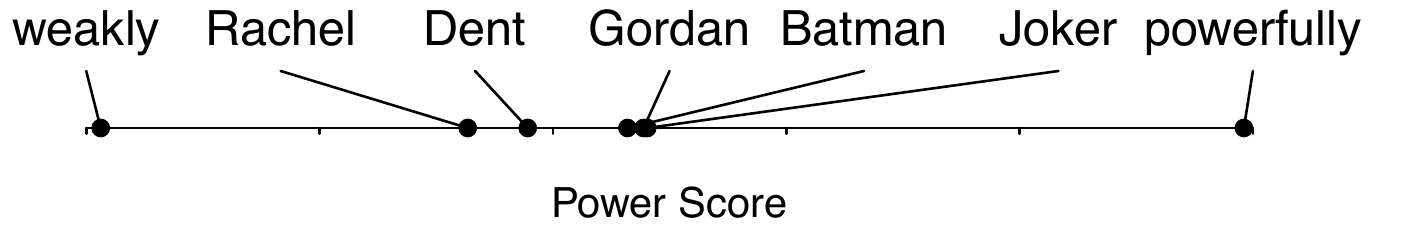} \\
    \vspace{10pt}
    \includegraphics[width=\linewidth]{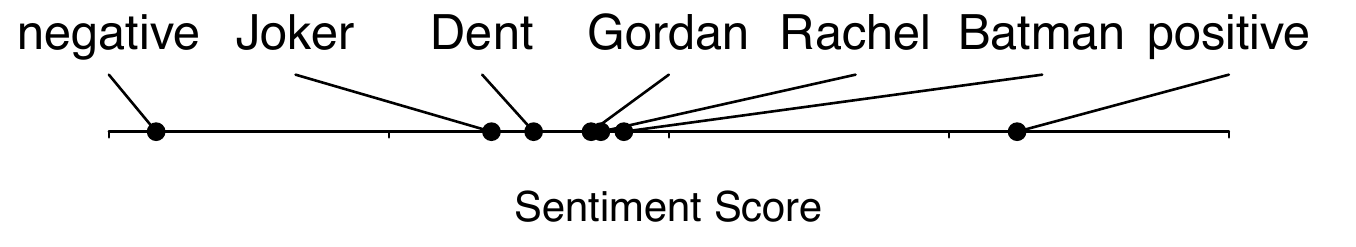} \\
    \vspace{10pt}
    \includegraphics[width=\linewidth]{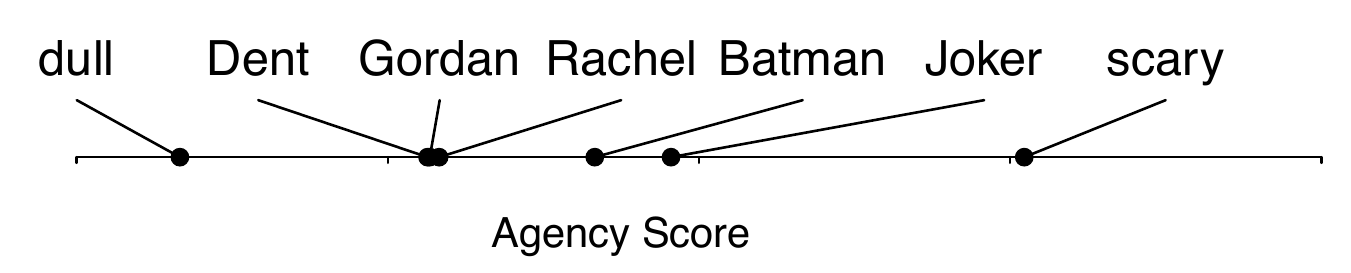}
    \caption{Power, sentiment, and agency scores for characters in \textit{The Dark Night} as learned through the regression model with ELMo embeddings. Scores generally align with character archetypes, i.e. the antagonist has  the lowest sentiment score.}
    \label{fig:darknight_reg}
\end{figure}

\subsection{Qualitative Document-level Analysis}
\label{sec:qual_entity}

Finally, we qualitatively analyze how well our method captures affect dimensions by analyzing single documents in detail.
We conduct this analysis in a domain where we expect entities to fulfill traditional power roles and where entity portrayals are known.
Following \citet{bamman2013learning}, we analyze the Wikipedia plot summary of the movie \textit{The Dark Knight},\footnote{\url{http://bit.ly/2XmhRDR}} focusing on Batman (protagonist),\footnote{We consider Batman/Bruce Wayne to be the same entity.} the Joker (antagonist), Jim Gordan (law enforcement officer, ally to Batman), Harvey Dent (ally to Batman who turns evil) and Rachel Dawes (primary love interest). To facilitate extracting example sentences, we score each instance of these entities in the narrative separately and average across instances to obtain an entity score for the document.\footnote{When we used this averaging metric in other evaluations, we found no significant change in results. Thus, in other scenarios, we compute scores over averaged embeddings, rather than averaging scores separately computed for each embedding to reduce computationally complexity.} To maximize our data by capturing every mention of an entity, we perform co-reference resolution by hand. Additionally, based on our results from \Tref{tab:forbes_power} as well as the use of Wikipedia data in training the ELMo model \cite{peters2018deep}, we use ELMo embeddings for our analysis.

\begin{figure}
    \centering
    \includegraphics[width=\linewidth]{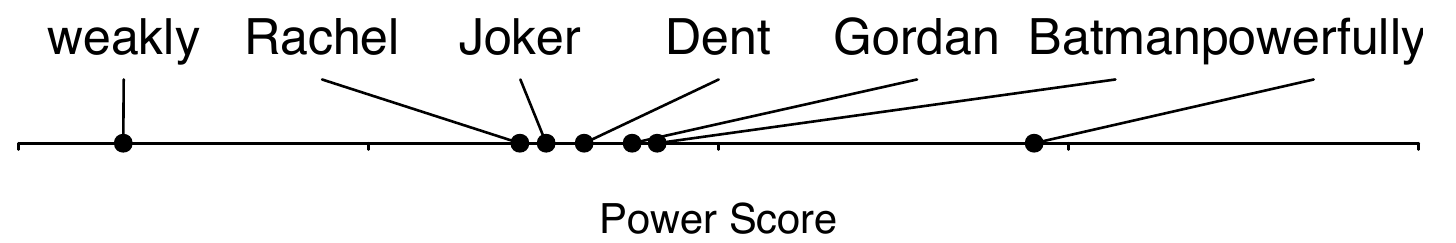} \\
    \vspace{10pt}
    \includegraphics[width=\linewidth]{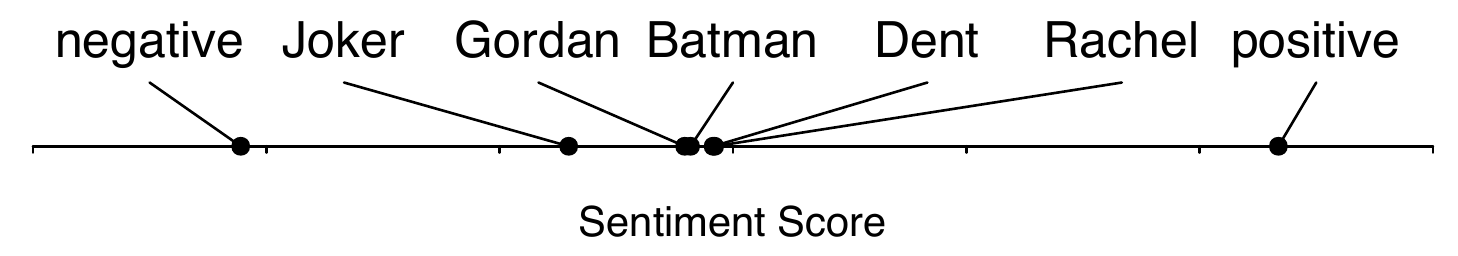} \\
    \vspace{10pt}
    \includegraphics[width=\linewidth]{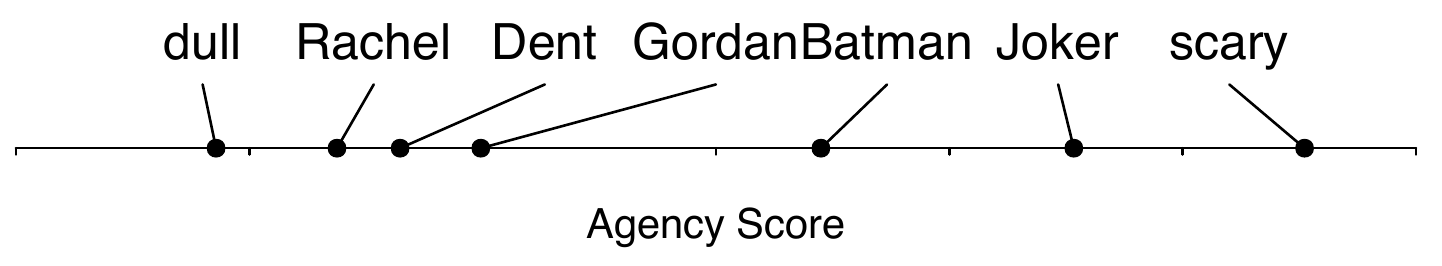}
    \caption{Power, sentiment, and agency scores for characters in \textit{The Dark Night }as learned through ASP with ELMo embeddings. These scores reflect the same patterns as the regression model with greater separation between characters.}
    \label{fig:darknight_pca}
\end{figure}

Figures \ref{fig:darknight_reg} and \ref{fig:darknight_pca}  show results. For reference, we show the entity scores as compared to one polar opposite pair identified by ASP. Both the regression model and ASP show similar patterns. Batman has high power, while Rachel has low power. Additionally, the Joker is associated with the most negative sentiment, but the highest agency. Throughout the plot summary, the movie progresses by the Joker taking an aggressive action and the other characters responding. We can see this dynamic reflected in the Joker's profile score, as a high-powered, high-agency, low-sentiment character, who is the primary plot-driver. In general, ASP shows a greater separation between characters than the regression model. We hypothesize that this occurs because ASP isolates the dimensions of interest, while the regression approach captures other confounds, such as that humans tend to be high agency entities.

Furthermore, because we score each instance separately, we can pinpoint particularly representative sentences. The sentence indicating the most positive sentiment for Batman is also the sentence that indicates the lowest sentiment for the Joker: ``Both the civilians and the prisoners refuse to kill each other, while Batman apprehends the Joker after a brief fight.''

An example sentence where the Joker is scored with particularly high power is: ``After announcing that Gotham City will be subject to his rule by nightfall, the Joker rigs two evacuating ferries with explosives.'' In contrast, a moment where Rachel is portrayed as particularly low-powered is: ``Both buildings explode, killing Rachel and disfiguring half of Dent's face.''

\begin{figure}
    \centering
    \includegraphics[width=\linewidth]{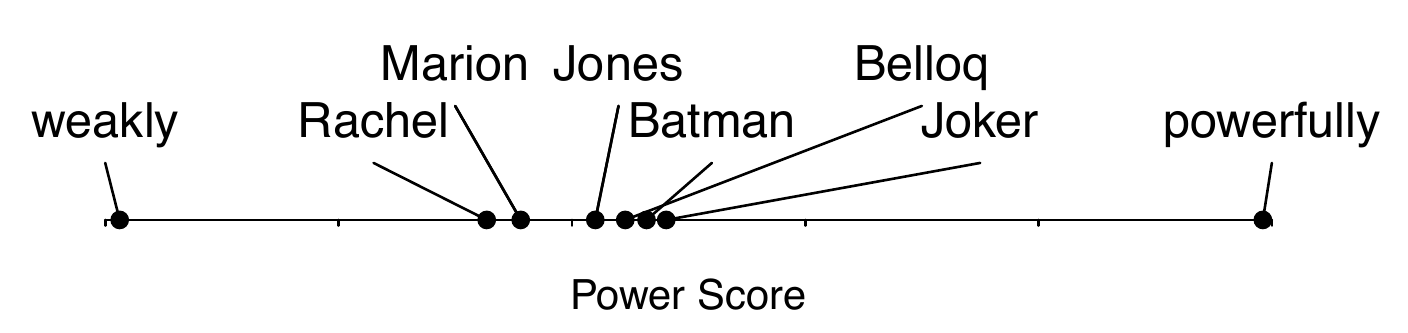}
    \caption{Power scores for characters in \textit{Raiders of the Lost Ark} and \textit{The Dark Night} as learned through the regression model with ELMo embeddings. Female characters have lower power scores than male characters.}
    \label{fig:indiana_power}
\end{figure}

\begin{figure}
    \centering
    \includegraphics[width=\linewidth]{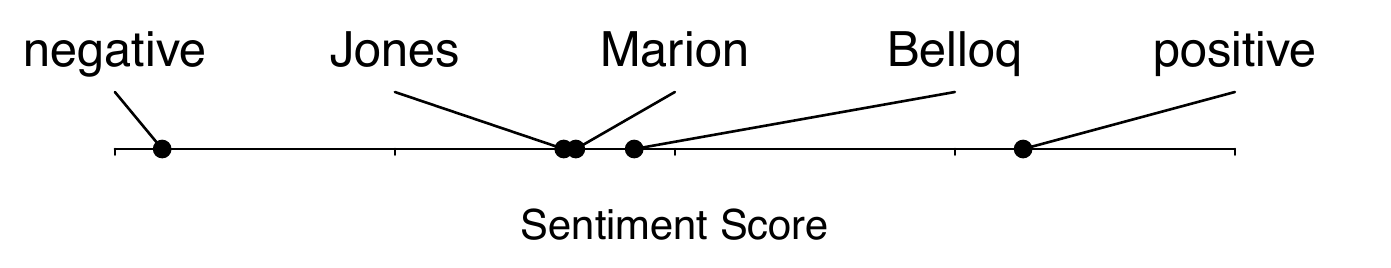}
    \caption{Sentiment scores for characters in \textit{Raiders of the Lost Ark} as learned through the regression model with ELMo embeddings. The antagonist is scored surprisingly positively.}
    \label{fig:indiana_sentiment}
\end{figure}

One of the advantages of the persona model in \citet{bamman2013learning} is the ability to cluster characters across stories, identifying roles like \textit{hero} and \textit{villain} more generally. We can similarly use our model to analyze characters across story lines. We show results using the regression model;  the ASP results (omitted) reveal the same patterns. In \Fref{fig:indiana_power}, we compare characters from the plot summary of \textit{Raiders of the Lost Ark} to the characters of \textit{The Dark Night}, specifically Indiana Jones (protagonist), Rene Belloq (antagonist) and Marion Ravenwood (love interest).\footnote{\url{http://bit.ly/30ZMhhj}} We can see a clear separation between the female love interests and the male protagonists and antagonists, thus identifying similar roles in the same way as a persona model. However, whereas the output of a persona model is distributions over personas and vocabulary, our system outputs scores along known dimensions of power, agency, and sentiment, which are easy to interpret and visualize. Furthermore, our approach is meaningful at the level of an individual document or sentence.

The affect scores in Indiana Jones reveal some of the limitations of our approach. \Fref{fig:indiana_sentiment} shows the sentiment scores for these characters. While Indiana Jones and Marion have similar sentiment scores, Belloq is portrayed surprisingly positively. In reading the plot summary, Belloq's role in the narrative is often not obvious through immediate context. While the Joker ``burns'' and ``rigs explosives'', Belloq ``arrives'' and ``performs a ceremonial opening''. The reader understands Belloq's role in the story through context in the broader story line, rather than context immediately surrounding mentions of Belloq. The sentence-level embeddings produced by ELMo do not capture the broader role of characters in narratives.

Finally, our model (as well as the persona model) does not specifically account for \emph{perspective}. For example, character deaths are often scored as a negative portrayal. Death may be a negative event, and often villains (i.e Belloq) die, allowing us to capture their role as negative characters. However, ``good'' characters also often die in stories, and in these cases, the reader tends to view the character positively (i.e. with sympathy). Our approach does not explicitly control for perspective, separating how an event may be negative from the perspective of a character but generate positive sentiment from the reader. The incorporation of connotations frames \cite{RashkinConnotationInvestigation}, in which annotations are along clearly defined perspectives,  may offer a way to improve our approach.

\section{Usage Example: Analysis of Gender Bias in Media}
\label{sec:usage}

\begin{figure}
    \centering
    \includegraphics[width=\linewidth]{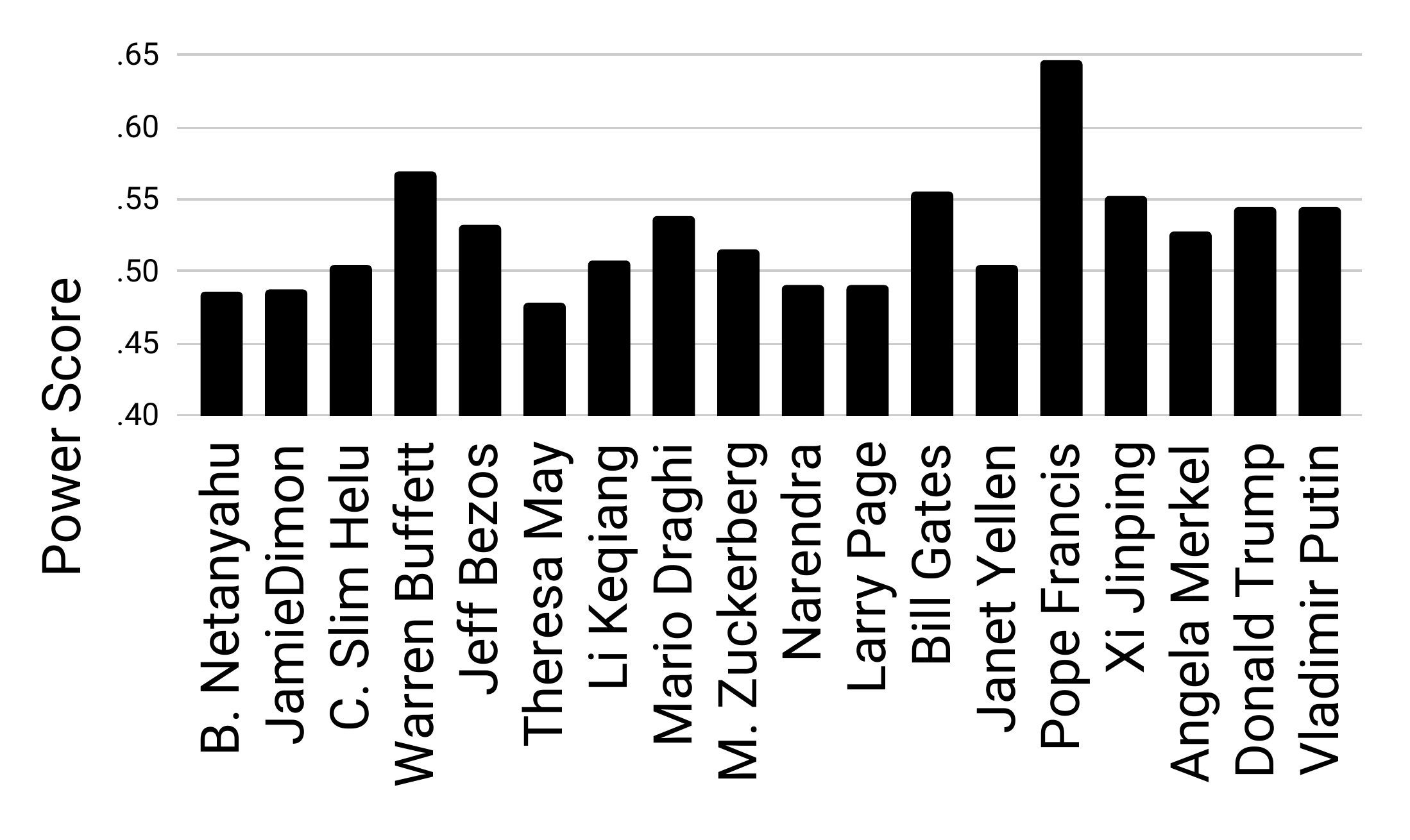}
    \includegraphics[width=\linewidth]{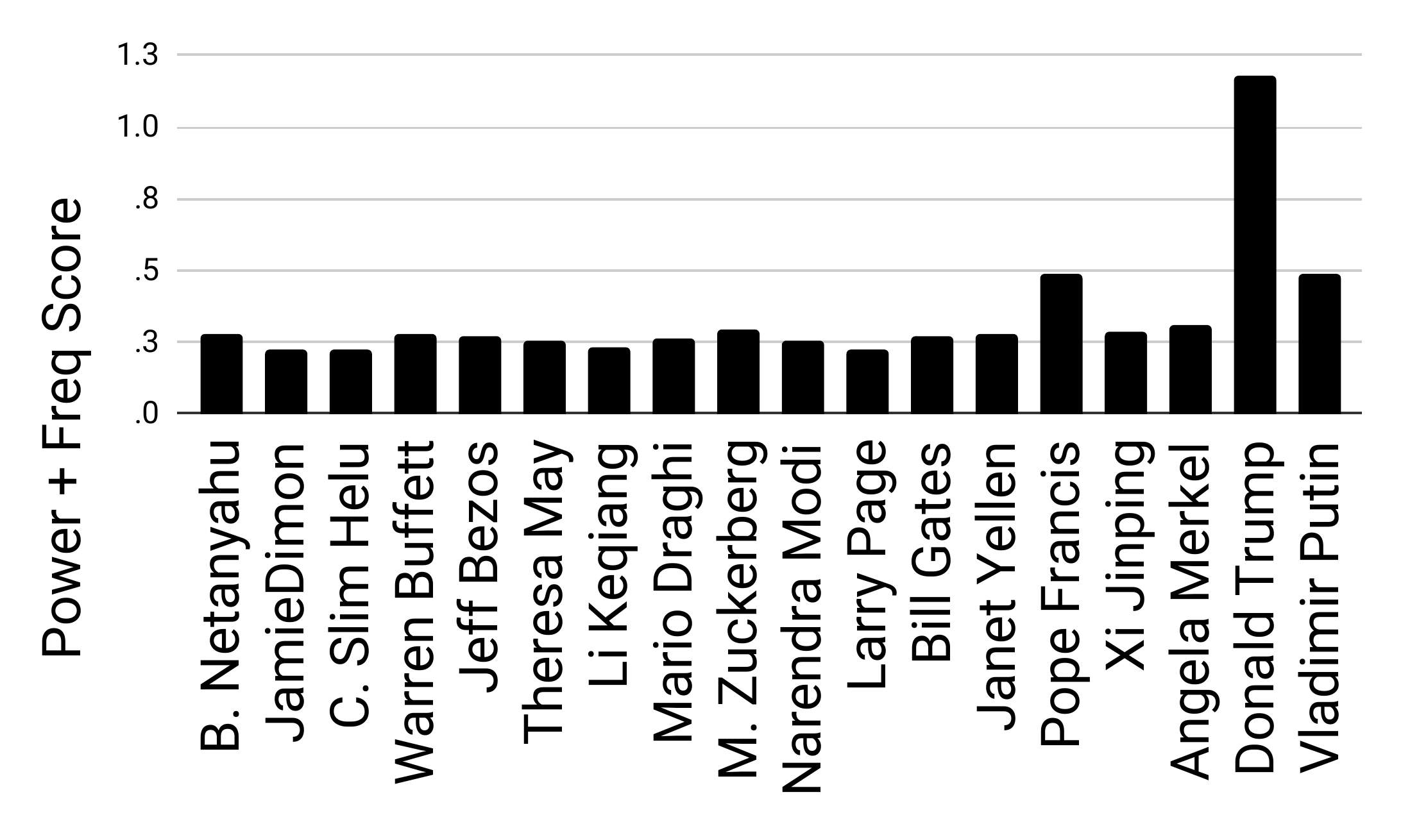}
    \caption{Power scores for people on the 2016 Forbes Magazine power list as learned through regression with ELMo embeddings, and through combined regression and frequency scores. Women are generally scored lower than similarly ranked men.\footnote{We omit two people who occurred infrequently in the corpus.}}
    \label{fig:forbes_power}
\end{figure}

In this section, we use our proposed methods to analyze how men and women are portrayed in the media, focusing on domains of interest in prior NLP work \cite{wagner2015s, fu2016tie}. We use the NOW corpus and regression with ELMo embeddings for analysis.\footnote{ASP results are nearly identical.}

First, we return to the example from \Sref{sec:quant_entity}, the list of most powerful people from Forbes Magazine in 2016.
\Fref{fig:forbes_power} shows the power scores ordered from least powerful to most powerful according to the Forbes list.
We show both the raw power scores computed by our model, as well as the regression power scores combined with frequency metric (as in \Tref{tab:forbes_power}).
In the raw scores, stand-out powerful people include businessman Warren Buffet and Pope Francis.
In contrast, the only 3 women, Theresa May, Janet Yellen, and Angela Merkel, are underscored as compared to similarly ranked men.
However, when we incorporate frequency, we do not see the same underscoring.
This result suggests that although these women are portrayed frequently in the media, they are typically described as less powerful than their actual role in society.\footnote{We note that the portrayals of other people with the same first names in the training data may have biased ELMo embeddings}
This finding is consistent with prior work on portrayals of women \cite{wagner2015s}.
The most striking difference after the incorporation of frequency scores is the boosted power score for Donald Trump, who is mentioned much more frequently than other entities.

\begin{figure}
    \centering
    \includegraphics[width=\linewidth]{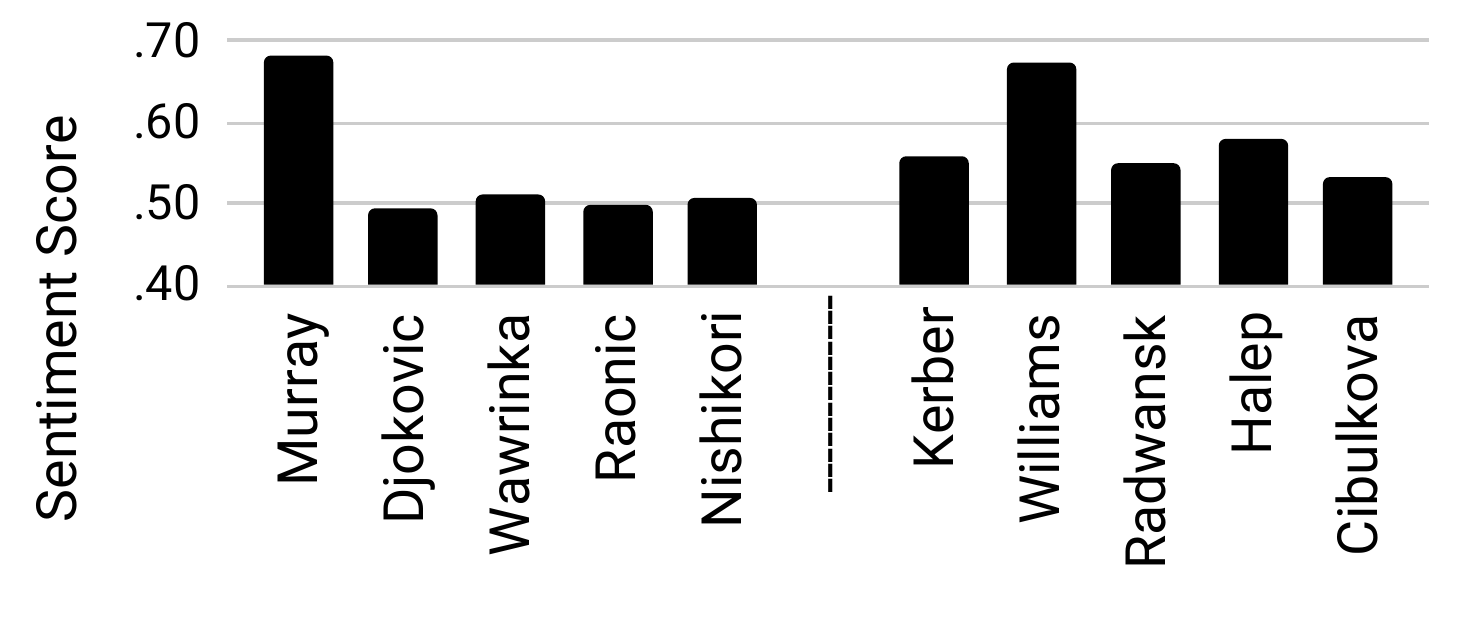}
        \includegraphics[width=\linewidth]{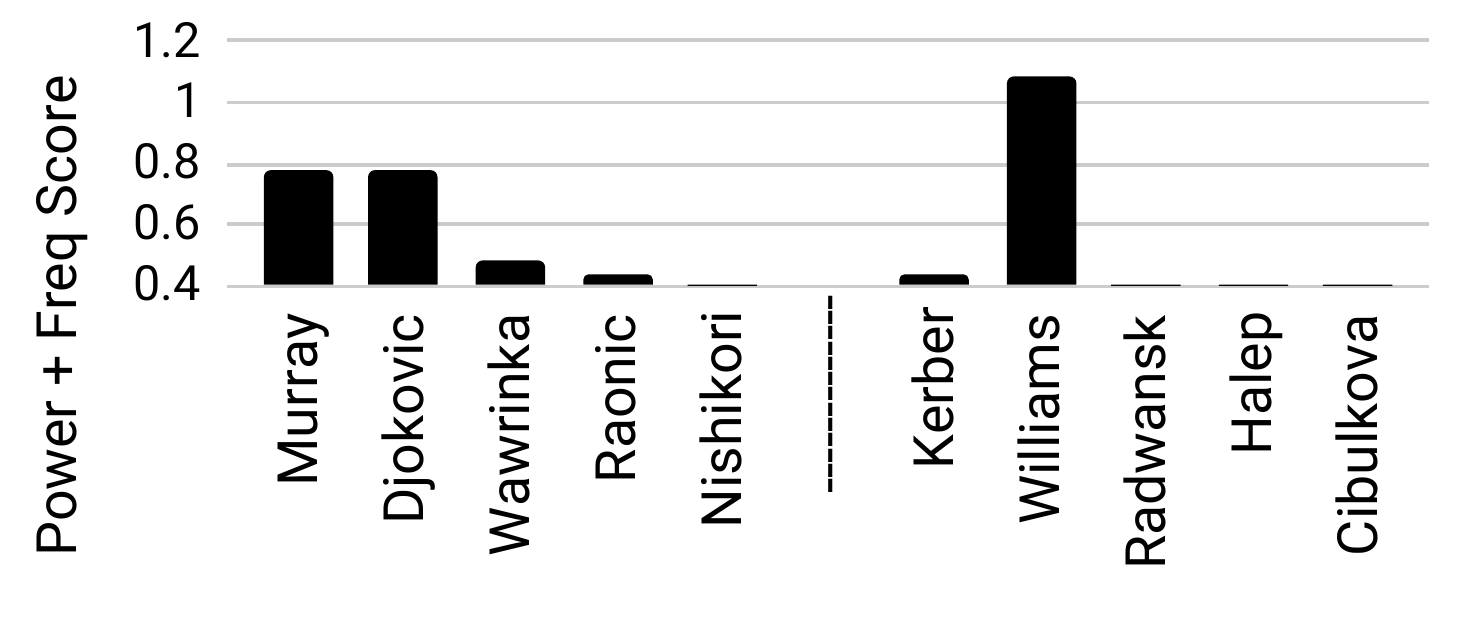}
    \caption{Sentiment and power scores for the top-ranked male (left) and female (right) tennis players in 2016 through regression with ELMo embeddings (power scores combine regression scores with frequency counts). Women are generally portrayed with lower power and higher sentiment.}
    \label{fig:tennis}
\end{figure}

In \Fref{fig:tennis}, we show the sentiment and power (combined regression + frequency) scores for the top-ranked male and female tennis players in 2016.
Prior work has shown bias in news coverage of male and female tennis players, specifically, that male players are typically asked questions more focused on the game than female players \cite{fu2016tie}.
Our analysis focuses on a different data set and coverage type---we examine general articles rather than post-match interviews.
As expected, popular players Serena Williams and Andy Murray have the highest sentiment scores and very high power scores. 
In contrast, Novak Djokovic, who has notoriously been less popular than his peers, has the lowest sentiment score, but the second highest power score (after Williams).
Additionally, female players are typically portrayed with more positive sentiment (female average score = 0.58; male average score = 0.54), whereas male players are portrayed with higher power (female average score = 0.52; male average score = 0.57).
However, the difference in power disappears when we remove frequency from the metric and use only the regression scores, suggesting that the difference occurs because male players are mentioned more frequently. 

\section{Related Work}

The most similar prior work to ours uses contextualized embeddings to map connotation frames (verb annotations) into power, agency, and sentiment scores for entities \cite{field2019}. In contrast, our method scores entities directly, allowing it to incorporate more information than just verb features and eliminating the need for dependency parsing. Furthermore, unlike the connotation frame annotations \cite{RashkinConnotationInvestigation, SapConnotationFilms}, the VAD lexicons used in this work were specifically motivated by social psychology literature on this topic, which influenced the annotation scheme \cite{vad-acl2018}. Our analysis in \Sref{sec:quant_entity} suggests that while \citet{field2019} works better for out-of-domain data, our proposed methods are able to obtain finer-grained and more accurate scores for in-domain data.

Prior to the proposed power, agency, and sentiment framework, initial approaches to person-centric analyses used graphical models to identify \emph{personas} in narratives \cite{bamman2013learning, card2016analyzing, iyyer2016feuding, chaturvedi2017unsupervised}, where personas are distributions over nouns, adjectives and verbs. These models allow for identifying roles in stories, such as Batman and Iron Man are both characters who ``shoot'', ``aim'', and ``overpower''. While this approach is useful for processing unstructured texts, personas are limited to distributions over a discrete vocabulary, and rely only on nouns, adjectives and verbs modifiers. In contrast, contextualized word embeddings have the power to capture all context in a sentence and provide more nuanced representations, especially considering non-contextualized embeddings have been shown to reflect biases in society \cite{garg2018word}. Furthermore, persona models can be difficult to interpret, whereas our analysis is grounded in concrete affect dimensions.

Other approaches that broadly address how people are portrayed use domain-specific features to target particular hypotheses. \citet{fast2016shirtless} analyze characters in fiction through crowd-sourced lexicons that target gender stereotypes. While useful for identifying bias, this method is limited to discrete modifiers and targeted lexicons do not necessarily generalize to other domains. \citet{wagner2015s} similarly use domain-specific knowledge to analyze coverage of men and women on Wikipedia, incorporating metadata like links between pages. Most affective NLP analyses of narratives focus on sentiment or specific stereotypes. Studies of power have largely been limited to a dialog setting (e.g. \citet{danescu2012echoes}, see \citet{prabhakaran2015social} for an overview), and almost no work has examined agency, with the exception of connotation frames.

Several recent works have evaluated the usefulness of pre-trained contextualized word embeddings in existing NLP tasks as well as through new benchmarks, designed to distill what type of information these models encode \cite{tenney2018what, goldberg2019assessing, liu2019}. These investigations focus on syntactic tasks, with semantic evaluations primarily limited to semantic role labeling. To the best of our knowledge, this is the first work to target affective dimensions in pre-trained contextualized word embeddings. Our findings are consistent with prior work suggesting that contextualized embeddings capture biases from training data \cite{zhao2019gender,keita19bias} and that these models perform best when trained on in-domain data \cite{alsentzer19}.

\section{Conclusions and Future Work}
We propose a method for incorporating contextualized word embeddings into entity-centric analyses, which has direct applications to numerous social science tasks.
Our results are  easy to interpret and readily generalize to a variety of research questions.
However, we further expose several limitations to this method, specifically that contextualized word embeddings are biased towards representations from their training data, which limits their usefulness in new domains.
While we explore masking target words as a possible solution to this problem, we find that masking significantly decreases performance.
We leave alternative solutions for future work, including training embeddings from scratch or fine-tuning on the target corpus (however, these ideas are only feasible with a large target corpus, and the need for fine-tuning reduces the usefulness of pre-trained embeddings).
Despite this limitation, we find that these models are expressive enough to analyze entity portrayals in in-domain data, allowing us to examine different portrayals of men and women. 

\section*{Acknowledgments}
 We gratefully thank anonymous reviewers, area chairs, Arnav Kumar, and Daniel Spokoyny. This material is based on work supported by the NSF GRFP under Grant No.~DGE1745016 and by Grant No.~IIS1812327 from the NSF. Any opinions, findings, and conclusions or recommendations expressed are those of the authors and do not necessarily reflect the views of the NSF.

\bibliography{acl2019}
\bibliographystyle{acl_natbib}

\appendix

\section{Appendix}
\label{sec:appendixA}
\begin{figure}[h!]
    \centering
     \includegraphics[width=0.3\linewidth]{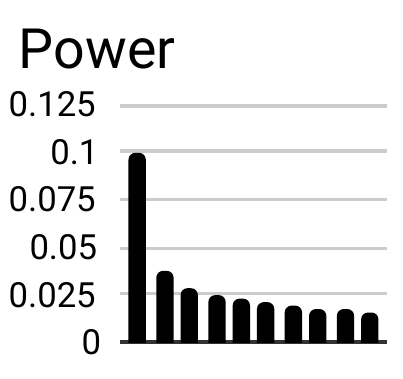}
    \includegraphics[width=0.3\linewidth]{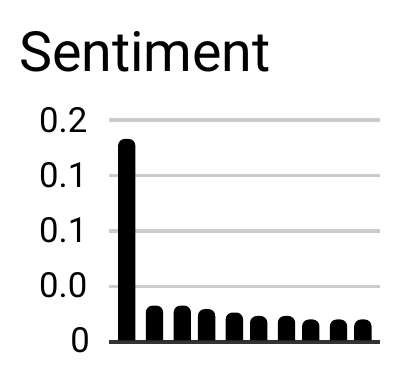}
    \includegraphics[width=0.3\linewidth]{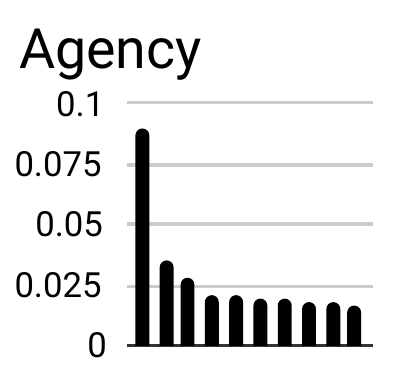}
    \caption{Percent of variance explained by the top 10 principle components for each affect dimension using ELMo embeddings. PCA was conducted on at least 100 embeddings per affect trait designed to have the greatest degree of variance along the dimension of the target affect trait}
    \label{fig:explained_variance}
\end{figure}

\section{Appendix}
\label{sec:appendix}
When using ELMo embeddings, we keep only the middle (second) ELMo layer, due to our preliminary investigations as well as prior work suggesting that this layer captures the most semantic information \cite{peters2018deep}. When constructing embeddings for multi-word entities we keep the embedding for the first word. 

The BERT model uses WordPiece embeddings \cite{wu2016google}, which can result in subword-level embeddings rather than word-level embeddings. In the case that a word is tokenized into subwords, we keep only the embedding for the first token in the word. We use the BERT Base Uncased model, and we use mean pooling to combine the 12 embedding layers into a single embedding with 768 dimensions.

We train hyper-parameters over the dev set, maximizing for Pearson correlation between the gold VAD annotations and the scores predicted by our models. We fix hyperparamters $|\mathcal{L}|$, $|\mathcal{H}|$, and $N$ as (400, 300, 200) for power, (900, 200, 100) for sentiment, and (400, 300, 200) for agency. In the regression model, we use an RBF kernel and fix $\alpha = 0.6$ and $\gamma = 1$. All embeddings are normalized to unit length.


\end{document}